\pdfoutput=1 

\documentclass[11pt]{article}
\usepackage{acl}
\usepackage{nert}
\usepackage{times,latexsym}
\usepackage{url}
\usepackage[T1]{fontenc}
\usepackage{physics} 
\usepackage{graphicx}
\usepackage{subcaption}
\usepackage{svg}

\title{Making Heads \emph{and} Tails of Models with Marginal Calibration\\ for Sparse Tagsets}

\author{Michael Kranzlein \\
  Georgetown University\\
  \textsmaller[.5]{\eml{mmk119@georgetown.edu}} \\\And
  Nelson F. Liu \\
  Stanford University\\
  \textsmaller[.5]{\eml{nfliu@cs.stanford.edu}} \\
  \\
  \And
  Nathan Schneider\\
  Georgetown University\\
  \textsmaller[.5]{\eml{nathan.schneider@georgetown.edu}}}

\date{}

\begin{document}
\maketitle

\begin{abstract}
For interpreting the behavior of a probabilistic model, it is useful to measure a model's \emph{calibration}---the extent to which the model produces reliable confidence scores.
We address the open problem of calibration for tagging models with \emph{sparse tagsets}, 
and recommend strategies to measure and reduce calibration error (CE) in such models. 
We show that several post-hoc \emph{re}calibration techniques all reduce calibration error across the marginal distribution for two existing sequence taggers.
Moreover, we propose \emph{tag frequency grouping} (TFG) as a way to measure calibration error in different frequency bands. 
Further, recalibrating each group separately promotes a more equitable reduction of calibration error across the tag frequency spectrum.

\end{abstract}

\section{Introduction}
An advantage of probabilistic models is that, in addition to providing a prediction, they also quantify \emph{uncertainty}. Knowing how certain a model is about a particular prediction can be crucial when using its output for downstream tasks or when weighing its trustworthiness. Of course, the probability estimate associated with a predicted output is an artifact of the model, and is subject to error---separate from the accuracy or error of the prediction itself.

By and large, NLP evaluations of multiclass classifiers and structured prediction models consider only the top \emph{prediction} for an input and how closely it matches the gold standard. Only in some studies is the \emph{probability} assigned to the prediction taken into account at all (e.g.~via a precision-recall curve).

A more comprehensive evaluation would examine whether the model's probabilities are \emph{well-calibrated}, i.e., whether they correlate well with empirical accuracy (such that $\approx\alpha$\% of predictions with probability close to $\alpha$ are in fact correct). \Citet{guo2017} showed that despite high accuracy, modern neural networks can still suffer from severe miscalibration. Fortunately, calibration error is not completely random, and can be corrected post hoc with a second model fit on development data (or even a separate recalibration set if available) as in several \emph{recalibration} techniques (\cref{sec:bg}).

In domains where NLP models help inform human decision-making (e.g., medicine), having a well-calibrated model is essential. Even in less critical domains, a well-calibrated model has potential to benefit rare instance discovery, pre-annotation, and self-training. In this paper we consider a structured prediction setting of particular relevance in NLP: tagging tasks with sparse tagsets---output spaces with a handful of high-frequency tags and many more rare tags.

Many linguistic phenomena follow power law distributions and thus feature a long tail of individually rare events, which, as we will show, makes it nontrivial to measure calibration error with existing methods, including marginal calibration error (MCE), which requires sufficient samples of each class to produce a reliable estimate \cite{kumar2019}. We evaluate two English sentence taggers\footnote{Data, code, and results are available at \url{https://github.com/nert-nlp/calibration_tfg}. Hyperparameters are described in \cref{sec:experiments}.} with closed sets of 100s of tags that disambiguate word tokens: a Combinatory Categorial Grammar (CCG) syntactic supertagger with 426 tags \citep{prange-21}, and a Lexical Semantic Recognition (LSR) tagger with 598 tags \citep{liu2021}.

Our main contributions are the following:
\begin{itemize}
    \item We posit that evaluation of calibration should go beyond a model's highest-confidence prediction, extending the arguments of \citet{nixon2020}, with a particular focus on sparse tagsets.
    
    \item We propose tag frequency grouping (TFG), a novel technique for evaluating and recalibrating groups of similarly frequent tags in a sparse tagging space.
    
    \item We introduce two new error metrics based on MCE suitable for tasks where insufficient data is available to apply MCE to all tags.
    
    \item We compare TFG and shared class-wise binning (SCW) on two sequence tagging tasks.

\end{itemize}

\section{Background}\label{sec:bg}

Calibration studies have two components: a recalibration technique and an evaluation metric. We use similar notation as \citet{kumar2019} to describe both.
That is, we assume a multiclass model $f:\mathcal{X}\rightarrow \mathcal{Y}$ that produces a real-valued score $f(X)_k \in [0, 1]$ for each class $k \in \mathcal{Y}$. In other words, for any input, the model gives $K = |\mathcal{Y}|$ scores. If these predictions are the output of a softmax function (as is typical for the last layer of neural networks), they will sum to 1 and can be interpreted as uncalibrated confidence scores across the distribution of possible classes or tags. The goal of recalibration is to make these confidence scores more reliable.

\subsection{Definition and Measurement}

There are several metrics for evaluating calibration error, including maximum calibration error \cite{naeini2015}, Brier Score \cite{brier1950}, calibration error (a term used widely in the literature, but here we refer to definition 2.1 in \citet{kumar2019}), and expected calibration error \cite{naeini2015}. We focus on the \emph{marginal calibration error} \cite{kumar2019}, which is a multiclass extension of CE.\footnote{\citet{kull2019} introduce a metric similar to MCE they call classwise-ECE.} MCE uses the $l_2$-norm to measure, for each class, ``the difference between the model's probability and the true probability of that class given the model's output'': $\textit{MCE}(f) = $

\begin{equation}\label{eq_mce}
    \sqrt{\frac{1}{K}\sum_{k=1}^{K}  \mathbb{E}\left[\left(f(X)_k - \mathbb{P}(Y = k \mid f(X)_k)\right)^2\right]}
\end{equation}

This metric is the root mean square error of measurements taken from $K$ binary recalibration models, where $\mathbb{P}$ is the true probability that the class is $k$ given $f(X)_k$, which is the model's predicted probability for class $k$ on input $X$. But one of the problems we quickly encounter with this definition (and similar measures of calibration error) is that with finite data, we cannot actually measure calibration error, since $f$ outputs values in a continuous range. In practice, this is overcome using binning schemes to estimate $\mathbb{P}(Y = k \mid f(X)_k)$. The range $[0, 1]$ is partitioned into bins; each score is placed in the appropriate bin; and error is estimated as the deviation between the average confidence of the bin and the proportion of positive labels in the bin (proportion of positive labels is equivalent to accuracy for top-label calibration).

\subsection{Recalibration Techniques}
We use three techniques for recalibration: histogram binning \cite{zadrozny2001}, isotonic regression \cite{zadrozny2002}, and scaling binning \cite{kumar2019}. All of these are \emph{post-hoc} techniques---they are applied after the model has been trained. In general, recalibration techniques fit into one of two categories: scaling or binning. Binning techniques quantize the interval of confidence scores and only output a fixed number of unique calibrated scores equal to the number of bins used for recalibration. Scaling techniques output continuous calibrated scores. Scaling techniques are generally better at reducing error, but because their output domain is continuous, the binning techniques used for evaluation are prone to underestimating true calibration error. \citet{kull2019} showed this with experiments on CIFAR-10 \cite{krizhevsky2009} and ImageNet \cite{russakovsky2015}.

\paragraph{Histogram Binning.}
Histogram binning is a popular recalibration technique that is simple and fast. The interval $[0, 1]$ is subdivided into $B$ subintervals using the confidence scores from the development set.\footnote{This is also referred to as a recalibration set in the literature, though they need not necessarily be disjoint.} The bin boundaries can be set such that each bin covers a fixed interval (fixed-width binning), or such that each bin includes the same number of data points \citep[adaptive binning;][]{nguyen2015}.

Using the boundaries for these $B$ bins, a confidence score from the test set is calibrated by finding the bin it belongs to and outputting the empirical proportion of positive labels among the development scores in that bin. This definition assumes a binary classification setting, but histogram binning can be extended to a multiclass scenario by building a one-vs.-rest model for each class, by using shared classwise binning \citep[SCW;][]{patel2021}, or by using TFG, described in \cref{sec:tfg}.

\paragraph{Isotonic Regression.}
Isotonic regression is a scaling technique that fits a non-decreasing piecewise linear function on the recalibration set by minimizing the square error subject to the non-decreasing constraint. It produces calibrated scores in a continuous range via linear interpolation.

\paragraph{Scaling Binning.}
Scaling techniques and binning techniques each have disadvantages. For example, histogram binning usually yields worse results than temperature scaling (another scaling technique), but its error measurement is reliable \cite{kumar2019}. Scaling binning combines the best of both approaches by first learning a scaling function. Uncalibrated scores are binned, and instead of outputting the proportion of positive labels (as in histogram binning), the calibrated score is the average output of the scaling function on the development scores in the bin. In our experiments with scaling binning, we use isotonic regression as the scaling function.

\subsection{Related Work}
\citet{zadrozny2002} initially proposed the one-vs.-rest approach for multiclass probabilities. \citet{kuleshov2015} recognize the sparsity problem and suggest reducing multiclass calibration of structured prediction to targeted ``events of interest'' and training a binary forecaster to learn calibrated probabilities of the event happening. This work is extended by \citet{jagannatha2020}, who treat a sequence of tags as a compositional model output and develop a forecaster based on gradient boosted decision trees. They achieve reductions in expected calibration error and a slight increase in model performance after reranking. Reranking refers to the process of normalizing calibrated scores and reordering them. With most recalibration techniques, it is rare for the ranking to be affected, and with some techniques like isotonic regression, the ranking of calibrated confidence scores will always match the uncalibrated ones.

\section{Designing and Evaluating Recalibration Models for Sparse Tagsets}\label{sec:eval}
The long tail of tags for CCG and LSR is of particular interest with respect to calibration. \Citet{kumar2019} point out that most studies of multiclass calibration focus primarily on \emph{top-label} calibration (reducing calibration error for only the top prediction out of the model for each input), also called top-1 or top-$k$ when looking at several of the model's top predictions. While \emph{top-label} scores are an important component of calibration, they don't tell the whole story, and we argue that the rest of the distribution (\emph{marginal} calibration) shouldn't be ignored. Recent works that address marginal calibration \cite{kumar2019, patel2021, nixon2020} make similar arguments but still tend to focus on balanced datasets like CIFAR-100, which contains 600 examples for each of 100 classes, or datasets with fewer tags like MNIST \cite{lecun1998}, MNIST Fashion \cite{xiao2017}, and CIFAR-10, which each have 10 classes.

In our analysis of marginal calibration, we study two long tails of distributions related to calibrating a sparse tagset: low confidence scores and low-frequency tags. We show how the standard \emph{one-vs.-rest} approach to multiclass calibration becomes infeasible as the size of the tagging space grows, and we provide specific recommendations for quantifying calibration error with sparse tagsets, where the lack of instances of rare tags poses unique challenges. 

Extending section~4 of \citet{nixon2020} with a particular focus on sparse tagsets, we now discuss the many design decisions that need to be made regarding multiclass calibration.

\subsection{Thresholding}

While we are interested in calibrating more of the distribution than is addressed with top-label calibration, it would be unwise to include all confidence scores. This is more an issue for evaluation than for recalibration. The justification for this decision is made clear in the distribution of the confidence scores and in prior work \cite{nixon2020}. We observe that more than 98\% of our two models' (which each have hundreds of possible tags) confidence scores are below 0.0001. Evaluating a recalibration model on all scores is likely to underestimate the error of the model, where the error on more likely output candidates will be washed out by excessively many near-zero scores that often have little error (particularly on a highly accurate model).

Instead, we select a threshold $t$ and if any scores are below this threshold, they are excluded from the recalibration and evaluation sets. For isotonic regression, including the scores below $t$ would have little effect as this scaling technique produces a piecewise function independent of any hyperparameter for the number of recalibration bins required for other techniques. However, if a threshold is not applied with binning techniques, many bins will contain only near-zero scores. For this reason (and consistency), we apply the threshold $t$ before both recalibration and evaluation for all techniques. Consequently, in our results we report calibration error on unnormalized scores, since thresholding excludes data and prevents us from obtaining calibrated scores for all tags in the distribution.

\subsection{Binning}

\paragraph{How should bin boundaries be determined?}
With a sparse tagset, it is even more important to avoid fixed-width binning, especially as the number of bins increases. Fixed-width binning will lead to significant imbalance, whereby the bins covering intervals of lowest and highest confidence scores will have many more items per bin, and the bins in the middle of the range will have very few items, causing high variance in estimates of calibration error. Thresholding does make the distribution less skewed, but many of the confidence scores in both of our datasets are low even after a threshold is applied. The alternative to fixed-width binning, adaptive binning \citep{nguyen2015}, puts the same number of items in each bin, leading to wider bins in the middle of the range, but guarantees each bin will have a sufficient number of data points for recalibration to overcome sampling error.

\paragraph{How to avoid too-small bins due to rare tags?}\label{par:eval}
Marginal calibration error as defined in \cref{eq_mce} treats each class as a binary recalibration problem and averages the error in each recalibration model that was estimated by binning. \citet{nixon2020} highlight that a finer-grained, per-class approach to evaluation analagous to MCE is ideal because it allows ``systematic differences in the calibration error between classes to be evaluated without washing each other out.'' In contrast to MCE, the top-label approach measures error only among the model's highest confidence score for each input (i.e.~the confidence score associated with the model's predicted label). This is done by binarizing the multiclass problem via one-hot labels. The top prediction of the model is selected and its gold label is taken to be 1 if that class is the true class and 0 otherwise. In this way, confidence scores for multiple tags can be evaluated together. This idea is key to how we modify MCE to evaluate our recalibration models. 

While MCE is the gold standard, it requires ample data in all tags in order to get a reliable measurement. With our sparse tagsets, measuring MCE separately for each tag is unfortunately infeasible, since we would not have enough samples in each bin. 
\Citet{nguyen2015}, for instance, recommend 200 samples per bin to reduce sampling error. In the literature, the floor for the number of bins used in evaluation is around 5. Assuming 5 bins at $\geq$200 samples each, that means creating a tag-specific recalibration model would require 1000 confidence scores. 

On its face, this is not a huge ask for marginal calibration with no thresholding, since having a recalibration set of 1000 tokens will produce 1000 confidence scores for each tag. But the number of near-zero confidence scores will increase as the tagset grows, and these near-zero scores are not as relevant to a discussion about calibration as actual candidate outputs from the model. For top-label calibration, it is possible to build a strong recalibration model, but in order to measure MCE for that model (with our assumption of 5 bins and at least 200 scores per bin), we would need at least 1000 tokens \emph{where each tag is predicted}. So the relative frequency of the rarest tag controls the total number of instances required for reliable binning (e.g., a tag occurring at a rate of 1\% would necessitate a recalibration set of 100,000 instances).

We experiment with two strategies to overcome this and derive a modified MCE metric. First, we extend the binarization approach of top-label error measurement to all labels, effectively creating a shared binning model for collective evaluation. This approach, shared classwise binning (SCW), was introduced by \citet{patel2021} for recalibration, but is extendable to evaluation. 
(We will introduce TFG, a generalization of SCW giving finer control over the sharing, in \cref{sec:tfg}.)

For SCW evaluation, we modify MCE and introduce shared marginal calibration error (SMCE). When operationalized with binning, we get \cref{eq_smce}. 

$\mathcal{D}$ contains the set of above-threshold confidence scores for all tokens and tags in the data.
In this equation, $\overline{q}_b$ is the average confidence score of the $b$-th bin and $\overline{p}_b$ is the average of the binary labels associated with each confidence score in the $b$-th bin. $N$ is the total number of confidence scores being recalibrated. $\textit{AdaBin}(\mathcal{D}, \beta)$ is our adaptive binning function that partitions the sorted confidence scores into bins of size $\beta$. 
A key difference between this metric and MCE is that scores for multiple tags are included in the square. 

\begin{equation}\label{eq_smce}
    \textit{SMCE}(\mathcal{D}, \beta) = \sqrt{\sum_{b \in \textit{AdaBin}(\mathcal{D}, \beta)} \frac{|b|}{N}(\overline{q}_b - \overline{p}_b)^2}
    \end{equation}

Using SCW for recalibration simply means learning a single recalibration model, pooling together all confidence scores from all tags.

\paragraph{How many bins should we use?}

We report results using 10 bins for recalibration and evaluation in our experiments, to ensure each bin has a sufficient number of datapoints.

\subsection{Tag Frequency Grouping}\label{sec:tfg}

\begin{figure*}
\centering
    \includegraphics[width=\textwidth]{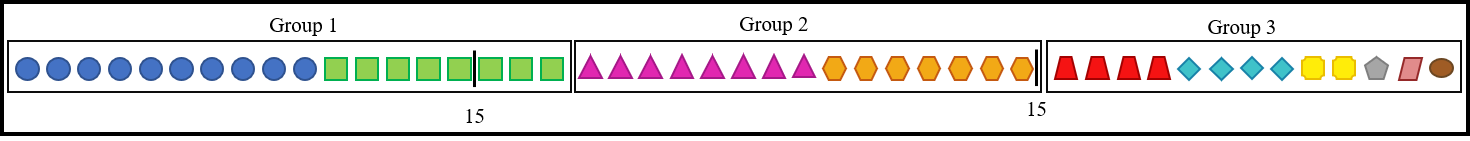}
    \caption{Illustration of tag frequency grouping (TFG) with 45 training instances and $G=3$ tag groups. Each shape represents a gold tag from the training data. Tags are sorted by frequency. Starting with the most frequent tag, groups are formed by iteratively adding all instances of a tag until the size of the group equals or exceeds the number of training instances divided by the number of groups. When TFG is used for recalibration (as opposed to just evaluation), a separate recalibration model is learned for each group.}
    \label{fig:tfg_example}
\end{figure*}

As we have explained, SCW solves the problem of rare tags by pooling all tags together when recalibrating or evaluating calibration error. 
But a concern is that this may be too coarse-grained: all tags are not necessarily created equal with respect to their calibration.
We therefore propose a new technique, TFG, to strike a balance between the two extremes of treating all tags together or independently with respect to calibration. TFG, like SCW, can be used for recalibration, evaluation, or both.

The intuition is simple. We often find that models are overconfident with tags seen frequently in the training data and underconfident with tags seen less frequently. Therefore, we hypothesize that \textbf{tags that are similarly frequent in the training data will be miscalibrated in similar ways}, and that by grouping together tags of similar frequencies and developing a separate recalibration model for each group, we can achieve improved results over SCW and calibrate tags that lack sufficient data for a class-specific recalibration model. The number of groups $G$ should be selected such that $G \ll K$, and in this paper, we report results where $G = 5$.\footnote{\Citet{patel2021} explored a similar idea in one of their experiments on digit recognition: digits with similar class priors were grouped together manually for recalibration. However, \citeauthor{patel2021} did not propose a general grouping technique, nor did they address large sparse tagsets as we do here.}

Choosing an optimal value of $G$ is tricky. As $G$ increases, the amount of recalibration data available for each group decreases, making each recalibration model less reliable. However, too low a value can lead to a reduced benefit over SCW with the loss of granularity (both in the recalibration models and in evaluation). Higher values for $G$ are likely suitable for larger datasets that still suffer from sparsity. However, if the dataset is sufficiently large and balanced, we recommend that independent recalibration models be created for each tag instead of using TFG or SCW.

In order to maximize generalization, we propose constructing tag groups based not on a model's output, but on the gold tag frequencies in the training data. The procedure is simple---sort the tags by descending frequency, and add the next most frequent tag to the group until the number of instances with gold tags in that group is greater than or equal to $1/G$. 

\Cref{fig:tfg_example} depicts a hypothetical example of TFG on a training set with 45 instances. Note that there's some overflow in the first group. This overflow can occur in any group except the last one, and in theory could lead to a worst-case scenario where the last group is much smaller than the others. In practice, this is unlikely to occur, but making sure all tag groups encompass a similar amount of training data is a good step to take prior to recalibration. 

SMCE (\cref{eq_smce}) can be adapted to grouped marginal calibration error (GMCE) for TFG by replacing $\mathcal{D}$, which contains confidence scores for all tags, with $\mathcal{G} \subseteq \mathcal{D}$, which contains confidence scores for one group (a subset of tags):

\begin{equation}\label{eq_gmce}
    \textit{GMCE}(\mathcal{G}, \beta) = \sqrt{\displaystyle\sum_{b \in \textit{AdaBin}(\mathcal{G}, \beta)} \frac{|b|}{N}(\overline{q}_b - \overline{p}_b)^2}
\end{equation}

\section{Experiments}

In our experiments, we develop recalibration models for two taggers with sparse tagsets and measure the improvement over the uncalibrated confidence scores with SMCE (overall error) and GMCE (per-group error).

\subsection{Taggers}

We consider two supervised tagging tasks trained and evaluated on different English datasets: CCG supertagging---a syntactic task with a large amount of training data and a high-accuracy model, and Lexical Semantic Recognition---a semantic task with less data and a lower-accuracy model.

\subsubsection{CCG Supertagging}
CCG is a lexicalized grammar formalism that is frequently used for syntactic and semantic parsing. CCG supertagging is the task of labeling each token with a complex, structured label that belies its function \citep{clark2002supertagging,bangalore2010supertagging}. \citet{bangalore1999supertagging} describe supertagging as ``almost parsing'', because a sequence of supertags maps a sentence to a small set of possible parses---the CCGBank \citep{hockenmaier2007ccgbank} dataset has over 1,200 unique CCG labels. By convention, the model is limited to predicting only tags that appeared at least 10 times in the training data, yielding 425 tags + the UNK tag. We use the non-constructive BERT-based \cite{devlin2019} model from \cite{prange-21} with its default hyperparameters. The tagger was trained on 927,497 tokens and obtained a dev accuracy of 96.1\%.

\subsubsection{Lexical Semantic Recognition}
LSR involves joint identification of multi-word expressions (MWEs), classification of lexical units, and disambiguation of coarse-grained supersenses and for noun, verb, preposition, and possessive expressions \cite{liu2021}. \Citet{liu2021} model this task as a sequence labeling problem using the STREUSLE dataset \citep{schneider-smith-2015-corpus,schneider-etal-2018-comprehensive}. For each token, they predict a tag with the conjunction of the token's MWE, lexcat, and supersense. Their model is also based on BERT, but it uses a conditional random field (CRF; \citealp{crf}) for decoding. We use a version of the model with no training or decoding constraints that has 598~tags and use its default hyperparameters. The tagger was trained on 44,801 tokens and obtained a dev accuracy of 81.1\%. To extract marginal distributions from the CRF, we use the Forward-Backward algorithm.

\subsection{Experimental Overview}\label{sec:experiments}
We use three techniques for recalibration: histogram binning, isotonic regression, and scaling binning with SCW and TFG. We use standard splits from the LSR and CCG datasets, fitting recalibration models on the development set and evaluating on the test set.

We exclude the one-vs.-rest recalibration setup from our experiments. The infeasibility of this approach with sparse tagsets is in fact one of the motivations for this paper. With SCW, there is one recalibration model per technique, and with TFG, there are $G$ independent recalibration models. We do not normalize the calibrated scores, since thresholding excludes many tags from the distribution on each sample.

For both grouping approaches and all three techniques, we evaluate tags in their respective frequency groups (GMCE) and collectively (SMCE). Evaluating with GMCE gives us more insight into which tags are miscalibrated (both before and after recalibration) and reduces exposure to cancellation effects among the different tags that could lead to an underestimation of error. Recall that an average of independent per-tag evaluations is the gold standard for mitigating these effects but is not possible due to how many tags lack sufficient representation in our datasets.

\paragraph{Summary of Hyperparameters and Reacalibration Model Design}
Following our explanations from \cref{sec:eval}, we made the following decisions for our models:

\begin{itemize}
    \item Apply a threshold and exclude all model predictions less than .01
    \item Use adaptive binning with 10~bins
    \item Use the $l_2$ norm for evaluation
    \item Evaluate error on unnormalized scores
    \item Set $G = 5$ for recalibration and evaluation with TFG
    
\end{itemize}

\section{Results and Discussion}

\begin{table*}[t]
\centering\small\setlength\tabcolsep{3.6pt}
\begin{tabular}{@{} l@{}r|c>{\smaller}c<{\smaller\%}
    | c>{\smaller}c<{\smaller\%}
    | c>{\smaller}c<{\smaller\%}
    || c>{\smaller}c<{\smaller\%}
    | c>{\smaller}c<{\smaller\%}
    | c>{\smaller}c<{\smaller\%} @{}}

\cline{3-8}\cline{9-14}
    & & \multicolumn{6}{c||}{\textbf{CCG}} & \multicolumn{6}{c|}{\textbf{LSR}} \\
\hline
\multicolumn{2}{c|}{\textbf{Recalibration}} & \multicolumn{2}{c|}{\textbf{All}} & \multicolumn{2}{c|}{\textbf{Group 1}} & \multicolumn{2}{c||}{\textbf{Group 5}} & \multicolumn{2}{c|}{\textbf{All}} & \multicolumn{2}{c|}{\textbf{Group 1}} & \multicolumn{2}{c}{\textbf{Group 5}}  \\
Method & $G$ & \textsc{smce} & \multicolumn{1}{c|}{$\Delta$} & \textsc{gmce} & \multicolumn{1}{c|}{$\Delta$} & \textsc{gmce} & \multicolumn{1}{c||}{$\Delta$} & \textsc{smce} & \multicolumn{1}{c|}{$\Delta$} & \textsc{gmce} & \multicolumn{1}{c|}{$\Delta$} & \textsc{gmce} & \multicolumn{1}{c}{$\Delta$} \\ \hline

\color{black} \textit{None} & \color{black} --- & \textit{.0167} & \multicolumn{1}{c|}{} & \textit{.0183} & \multicolumn{1}{c|}{} & \textit{.0396} & \multicolumn{1}{c||}{} & \textit{.0330} & \multicolumn{1}{c|}{} & \textit{.0356} & \multicolumn{1}{c|}{} & \textit{.0553} & \multicolumn{1}{c}{}\\[5pt]

\color{blue} Scaling Binning & \color{blue} 1 & .0065 & $-$61.0 & .0225 & 22.89 & .0390 & \hphantom{0}$-$1.47 & .0144 & $-$56.36 & .0105 & $-$70.48 & .0534 & \hphantom{0}$-$3.48\\
\color{dkgreen} Scaling Binning & \color{dkgreen} 5 & .0019 & $-$88.87 & .0049 & $-$73.09 & .0114 & $-$71.19 & .0159 & $-$51.83 & .0269 & $-$24.42 & .0153 & $-$72.27\\[5pt]

\color{blue} Isotonic Regres. & \color{blue} 1 & .0024 & $-$85.54 & .0226 & 23.72 & .0276 & $-$30.35 & .0130 & $-$60.74 & .0161 & $-$54.95 & .0374 & $-$32.49\\
\color{dkgreen} Isotonic Regres. & \color{dkgreen} 5 & .0032 & $-$80.93 & .0230 & 25.58 & .0228 & $-$42.37 & .0124 & $-$62.57 & .0132 & $-$62.93 & .0142 & $-$74.34\\[5pt]

\color{blue} Histogram Bin. & \color{blue} 1 & .0047 & $-$72.06 & .0286 & 56.49 & .0422 & \hphantom{$-$0}6.54 & .0086 & $-$73.94 & .0298 & $-$16.4 & .0409 & $-$26.15\\
\color{dkgreen} Histogram Bin. & \color{dkgreen} 5 & .0028 & $-$83.42 & .0254 & 38.69 & .0218 & $-$44.93 & .0110 & $-$66.76 & .0222 & $-$37.78 & .0145 & $-$73.71\\[5pt]\hline

\multicolumn{2}{c}{$N$} & \multicolumn{2}{c}{72,373} & \multicolumn{2}{c}{12,873} & \multicolumn{2}{c||}{15,854} & \multicolumn{2}{c}{15,933} & \multicolumn{2}{c}{2,854} & \multicolumn{2}{c}{3,020} \\
\multicolumn{2}{c}{Tag types} & \multicolumn{2}{c}{415} & \multicolumn{2}{c}{1} & \multicolumn{2}{c||}{382} & \multicolumn{2}{c}{377} & \multicolumn{2}{c}{3} & \multicolumn{2}{c}{302} \\
\multicolumn{2}{c}{Tag freq in train} & \multicolumn{2}{c}{} & \multicolumn{2}{c}{[22.2\%,22.2\%]} & \multicolumn{2}{c||}{[.0\%,0.4\%]} & \multicolumn{2}{c}{} & \multicolumn{2}{c}{[7.1\%,10.3\%]} & \multicolumn{2}{c}{[.0\%,.1\%]} \\
\multicolumn{2}{c}{Tokens} & \multicolumn{2}{c}{55,371} & \multicolumn{2}{c}{12,873} & \multicolumn{2}{c||}{9,167} & \multicolumn{2}{c}{5,381} & \multicolumn{2}{c}{2,739} & \multicolumn{2}{c}{1,716}
\end{tabular}
\caption{Marginal calibration error (measured with SMCE and GMCE) before and after recalibration with different techniques on two tasks: Combinatory Categorial Grammar (CCG) supertagging and Lexical Semantic Recognition (LSR). These data are visualized in \cref{fig:results_5_groups}. SMCE indicates shared marginal calibration error, and GMCE indicates grouped marginal calibration error (see \cref{par:eval}); $\Delta$ refers to the relative change over the original model (lower is better). 5 groups are used for tag frequency--based evaluation; only the highest-frequency tags (Group~1) and lowest-frequency tags (Group~5) are shown. The TFG conditions use the same 5~groups for separate recalibration models, while the SCW conditions ($G = 1$) use multiple groups only for evaluation.\finalversion{(See \nss{appendix ref}\mk{do we still want this?} for top-1 calibration results.)}}
\label{tab:results}
\end{table*}

\begin{figure}[h!]
    \centering
    \includegraphics[width=.4\textwidth]{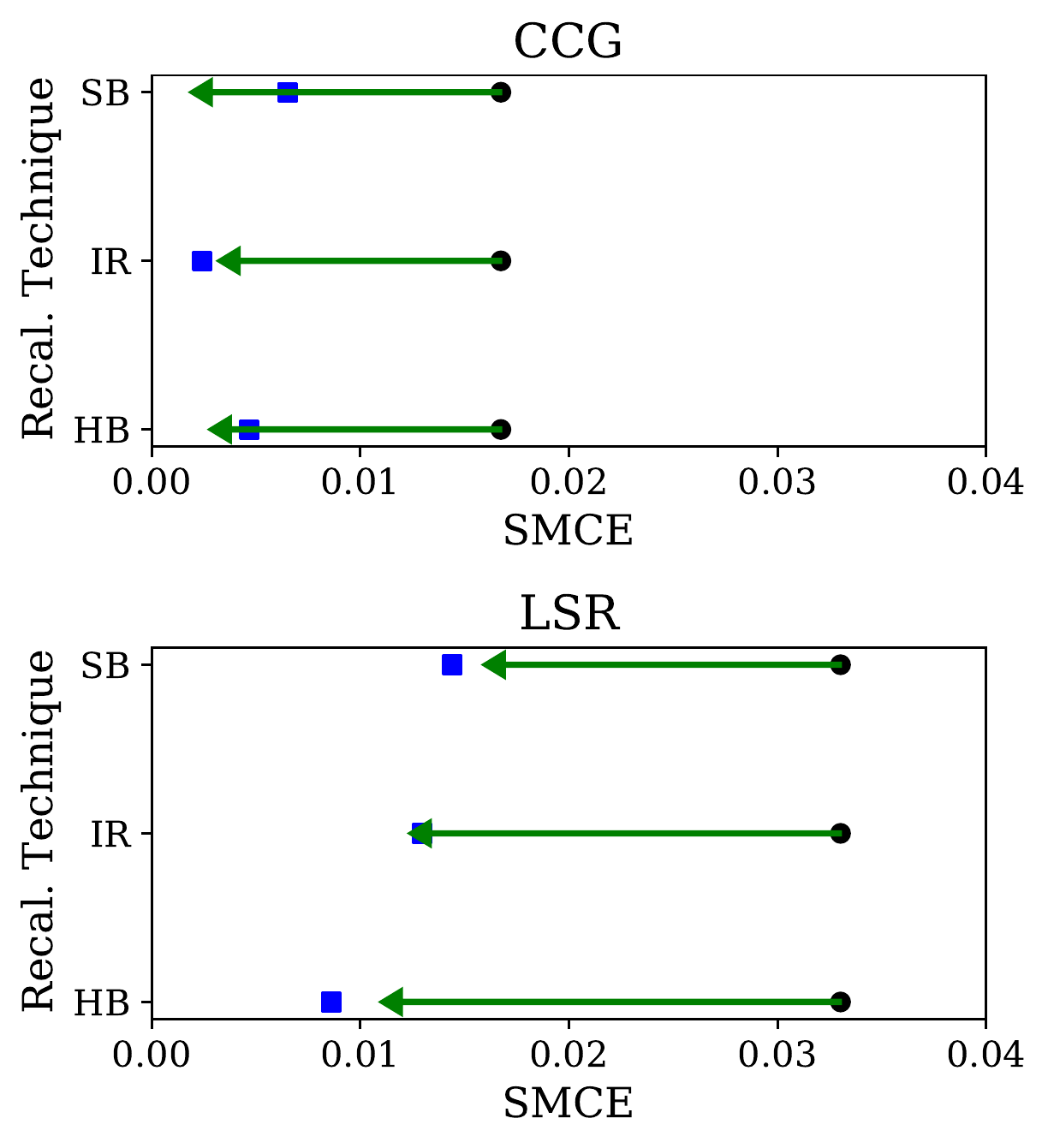}
    \caption{Evaluation of recalibration techniques, TFG, and SCW using SMCE. Techniques include histogram binning (HB), isotonic regression (IR), and scaling binning (SB) using isotonic regression as the scaling function. Black circles show initial calibration error; green arrows pointing to the left show reductions in calibration error after recalibration with TFG ($G=5$); and blue squares show calibration error after recalibration with SCW. Lower SMCE is better.}
    \label{fig:results_collective}
\end{figure}

\begin{figure*}
    \centering
    \includegraphics[width=\textwidth]{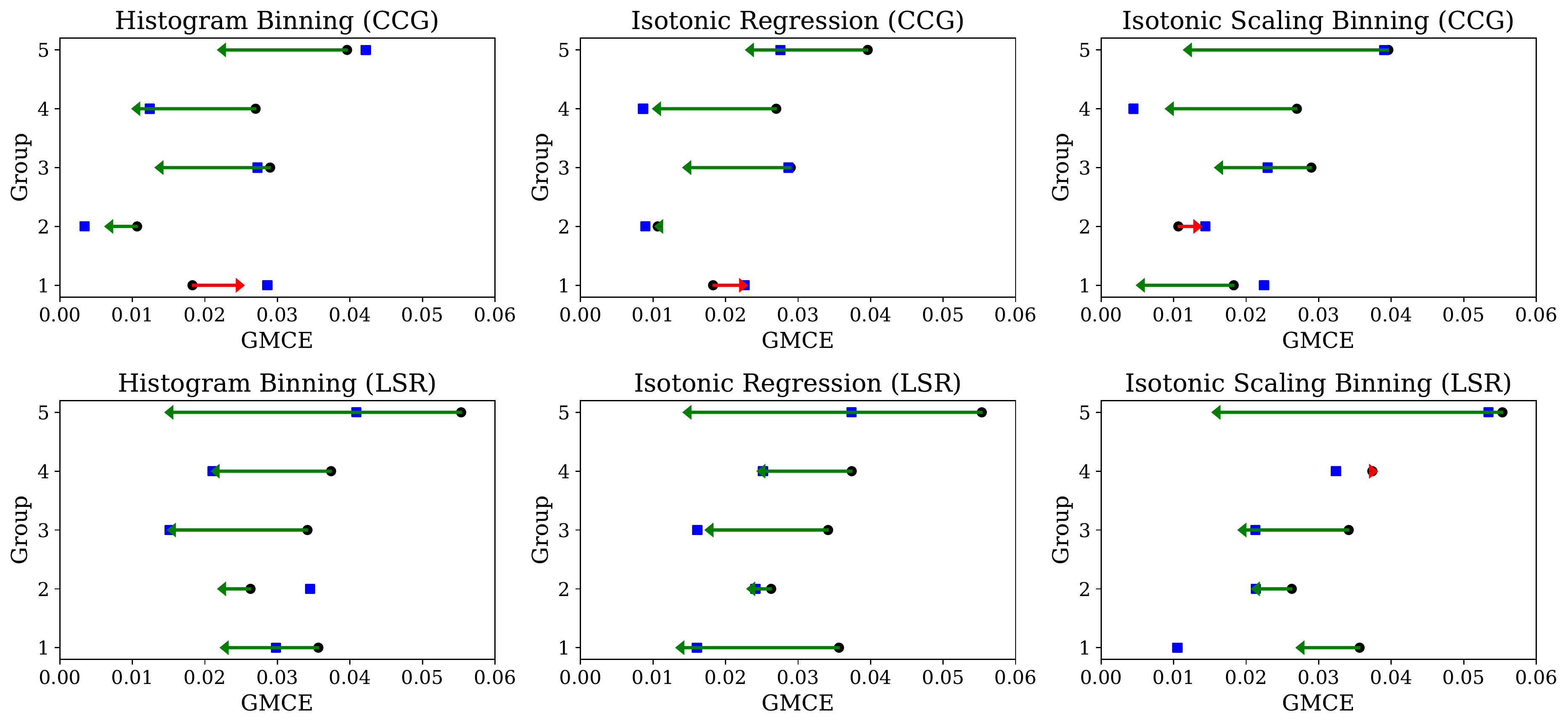}
    \caption{Evaluation of recalibration techniques, TFG, and SCW on 5 groups of tags. Colors and shapes are the same as \cref{fig:results_collective}, but red arrows pointing to the right indicate an increase in calibration error. Group 1 includes the most frequent tags (as represented in the training data), and Group 5 includes the rarest tags. Lower GMCE is better.}
    \label{fig:results_5_groups}
\end{figure*}

Our experimental results are visually summarized in \cref{fig:results_collective,fig:results_5_groups}. \Cref{tab:results} provides total error across the marginal distribution as well as the error in the most frequent tags and least 
frequent tags. 

Overall, our models for both datasets benefit from recalibration and see substantial reductions in calibration error with SCW, TFG, and all recalibration techniques. Relative to the CCG model, the LSR model has higher absolute error, and we see greater relative improvements from recalibration. The recalibrated CCG model has the lowest \emph{absolute} error.

\paragraph{How do post-hoc techniques compare?}
We evaluated three recalibration techniques in our experiments: histogram binning, isotonic regression, and scaling binning. When measuring calibration error collectively in \cref{fig:results_collective}, we noticed similar performance. 
Breaking the error down by tag group in \cref{fig:results_5_groups} gives us more insights about how our recalibration techniques affect tags of different frequency. All of the techniques achieve similar performance, though isotonic regression has the fewest outliers, with only one situation—Group 1 for CCG—where calibration error gets worse.

 Both binning techniques, and in particular histogram binning, are susceptible to making things worse in some cases. This happens more with the CCG tagger, which was fairly well calibrated to begin with. It is more accurate than the LSR tagger and has high average confidence in its output, with relatively few confidence scores near 50\%. 

 In recalibration with binning methods, this makes CCG more prone to unlucky wide bin boundaries (which are more likely to have high error). Using more bins for recalibration could help mitigate this problem; we used 10 bins for both models for parity in comparisons. While isotonic regression appears the most reliable, it does not have the same quantifiable error bounds as scaling binning \cite{kumar2019}, which should be taken into account when choosing a recalibration technique.

\paragraph{How do groups compare?}
For both datasets, Group 2 has the lowest initial calibration error, and it sees some of the smallest changes after recalibration. These tags are still frequent in the training data,
but less so than the tags in Group 1. Group 5, which contains the rarest tags, has the highest calibration error and sees the biggest improvements. 

The statistics at the bottom of \cref{tab:results} show us how unbalanced our tagsets truly are. $N$ shows the number of confidence scores that exceed the threshold. Then there is the number of tags represented in each group and the minimum and maximum training frequencies of the tags in each group. ``Tokens'' shows the number of tokens with any score above the threshold in each group. Remarkably, Group 1 for CCG has just 1 tag, \texttt{N}, that makes up more than 22\% of the gold-labeled data in the training set, whereas the most frequent tag for Group 5 is just .4\% of the gold-labeled data.

\paragraph{How does TFG compare to SCW for recalibration?}
TFG performs drastically better than SCW on the rarest tags. In most cases for other tag groups, the TFG results and SCW results are close. Only with scaling binning on LSR for Group 1 does SCW outperform TFG by a wide margin. This may be the result of a lucky bin boundary, as SCW does worse than TFG with histogram binning for LSR. 

Groups 1 and 2 are interesting for CCG. With SCW, all three techniques \emph{increased} calibration error for Group 1. With TFG, histogram binning appears to sacrifice performance on Group 1 for the benefit of Group 2, and scaling binning does the opposite.

TFG yields strong improvements in all other tag groups for CCG, whereas SCW does not. The only other case where TFG slightly increases calibration error is Group 4 for LSR with scaling binning.

Our results suggest that when used for recalibration, TFG yields overall improvements in calibration error that are similar to or better than SCW, especially on less frequent tags. For datasets where SCW might outperform TFG, we can still recommend TFG for evaluation of models with sparse tagsets via GMCE, since GMCE provides more information about which tags suffer from the greatest miscalibration.

\section{Conclusion}
We examined the challenges of evaluating and reducing calibration error with sparse tagsets.
In particular, we introduced TFG to offer more control over how tags are pooled together given that some are too infrequent to be recalibrated\slash evaluated independently. We showed that SCW and TFG are easily extensible from recalibration to the evaluation setting with the SMCE and GMCE metrics, and that GMCE gives more specific insight into where in a tag distribution the most calibration error exists and where it can be reduced. On one semantic task and one syntactic task, we found substantial improvement in calibration error for the head and tail of the tag distribution.

Opportunities for further research include devising methods for choosing and evaluating the optimal value for $G$ and considering normalizing scores despite the elimination of scores below the threshold. While the recalibrated model would be unable to assign any confidence to tags excluded by thresholding, this effect may be minimal, and it could lead to improved interpretability since the distribution would sum to 1.

It may also be worth relying not just on frequency but incorporating the structure of each tag into the grouping process. LSR and CCG tags, for example, are compositional, and could be grouped based on subtag. Testing whether TFG has benefits for more balanced tagsets is another opportunity.

\section*{Acknowledgements}
We are grateful to anonymous reviewers as well as members of the NERT lab for their feedback on this work.
This research was supported in part by NSF award IIS-1812778 and grant 2016375 from the United States--Israel Binational Science Foundation (BSF), Jerusalem, Israel.

\bibliography{calibration}
\bibliographystyle{acl_natbib}

\end{document}